\documentclass{article}

\usepackage[preprint]{neurips_2025}

\usepackage[utf8]{inputenc} 
\usepackage[T1]{fontenc}    
\usepackage{hyperref}       
\usepackage{url}            
\usepackage{booktabs}       
\usepackage{amsfonts}       
\usepackage{nicefrac}       
\usepackage{microtype}      
\usepackage{xcolor}         

\usepackage{subcaption}
\usepackage{graphicx}
\usepackage{amsmath}
\usepackage{amssymb}
\usepackage{adjustbox}
\usepackage{makecell}
\usepackage{multirow}
\usepackage{bbding}
\usepackage{color, colortbl}
\usepackage{enumitem}

\usepackage{amsthm}

\usepackage{circledsteps}

\usepackage{pifont}

\usepackage{algorithm}
\usepackage{algorithmicx}
\usepackage{algpseudocode}

\title{DuQuant++: Fine-grained Rotation \\ Enhances Microscaling FP4 Quantization}

\author{Haokun Lin$^{*\ 1,6}$, Xinle Jia$^{*\ 2}$, 
Haobo Xu$^{3}$, 
Bingchen Yao$^{4}$,
Xianglong Guo$^{1}$
\vspace{0.1cm},\\
\textbf{Yichen Wu$^{5,6}$, Zhichao Lu$^6$, Ying Wei$^4$,  Qingfu Zhang$^6$,
Zhenan Sun$^{1}$ }\vspace{0.2cm}\\
  $^*$Equal Contribution\vspace{0.2cm} \\
    $^1$ CASIA \quad
    $^2$ NJU \quad
    $^3$ THU \quad
    $^4$ ZJU \quad
    $^5$ Harvard \quad
    $^6$ CityU \\
    \vspace{-0.3cm}
}

\begin{document}

\maketitle

\begin{abstract}
The MXFP4 microscaling format, which partitions tensors into blocks of 32 elements sharing an E8M0 scaling factor, has emerged as a promising substrate for efficient LLM inference, backed by native hardware support on NVIDIA Blackwell Tensor Cores.
However, activation outliers pose a unique challenge under this format: a single outlier inflates the shared block scale, compressing the effective dynamic range of the remaining elements and causing significant quantization error.
Existing rotation-based remedies, including randomized Hadamard and learnable rotations, are \emph{data-agnostic} and therefore unable to specifically target the channels where outliers concentrate.
We propose \textbf{DuQuant++}, which adapts the outlier-aware fine-grained rotation of DuQuant to the MXFP4 format by aligning the rotation block size with the microscaling group size ($B{=}32$).
Because each MXFP4 group possesses an independent scaling factor, the cross-block variance issue that necessitates dual rotations and a zigzag permutation in the original DuQuant becomes irrelevant, enabling DuQuant++ to replace the entire pipeline with a single outlier-aware rotation, which halves the online rotation cost while simultaneously smoothing the weight distribution.
Extensive experiments on the LLaMA-3 family under MXFP4 W4A4 quantization show that DuQuant++ consistently achieves state-of-the-art performance.
Our code is available at \url{https://github.com/Hsu1023/DuQuant-v2}.
\end{abstract}

\section{Introduction}

Large language models (LLMs) have achieved remarkable performance across a wide range of tasks, yet their deployment is increasingly constrained by the substantial memory footprint and computational cost during inference~\citep{dubey2024llama3,zhou2025scale,xu2026prune}.
Post-training quantization (PTQ) has emerged as one of the most practical solutions, enabling model compression with only a small calibration set and no retraining~\citep{frantar2022gptq,xiao2023smoothquant,xie2025automated,zhang2026quantvla}.
While early efforts focused on integer formats, progressing from INT8~\citep{dettmers2022llm_int8} to aggressive 4-bit weight-activation (W4A4) settings~\citep{ashkboos2024quarot,lin2024duquant,sun2024flatquant}, the hardware landscape is shifting toward \emph{floating-point microscaling formats} that promise higher numerical fidelity at comparable bit budgets.

Among these emerging formats, MXFP4~\citep{rouhani2023ocp} partitions tensors into small blocks of 32 elements and assigns each block a shared scaling factor encoded in the E8M0 format. 
With native hardware support from NVIDIA Blackwell Tensor Cores~\citep{tirumala2024nvidia}, MXFP4 offers an attractive balance between compression ratio and hardware efficiency for LLM inference.
However, this block-wise design introduces a unique challenge: because all 32 elements within a microscaling group share a single scaling factor, any outlier in the group directly inflates the shared scale, compressing the effective dynamic range available for the remaining elements and significantly increasing quantization error.
This problem is particularly severe in LLM activations, where both normal outliers~\citep{xiao2023smoothquant} and massive outliers~\citep{sun2024massive,liu2024intactkv} are prevalent at certain positions such as the down projection input.

Several recent works have attempted to extend rotation-based quantization technique. 
MR-GPTQ~\citep{egiazarian2025mr-gptq} and BRQ~\citep{shao2025brq} adopt block-wise randomized Hadamard rotations to spread outlier energy within each microscaling group.
QuaRot~\citep{ashkboos2024quarot} applies a global Hadamard rotation that mixes all channels before quantization.
FlatQuant~\citep{sun2024flatquant} learns end-to-end rotation matrices optimized for quantized inference.
Despite their contributions, these methods share a common limitation: the rotation matrices are \emph{data-agnostic}, which are either random or learned without awareness of the actual outlier structure.
As a result, they treat all feature dimensions equally, missing the opportunity to specifically target the channels where outliers are most concentrated.
Moreover, global rotations destroy the block-wise independence of MXFP4 groups, while the computational overhead of learnable rotations can be non-trivial.

In this work, we propose \textbf{DuQuant++}, which adapts the outlier-aware fine-grained rotation from DuQuant~\citep{lin2024duquant} to the MXFP4 microscaling format.
The key idea is to \emph{align the rotation block size with the MXFP4 group size} ($B = 32$), so that each rotation block operates precisely within one microscaling group.
This alignment brings a crucial simplification: since each MXFP4 group has its own independent scaling factor, the quantization error of one group does not affect another, and there is no shared global scale that could be inflated by a single group's outlier.
Consequently, the cross-block variance issue that necessitates the zigzag permutation and second rotation in the original DuQuant~\citep{lin2024duquant} becomes irrelevant, allowing DuQuant++ to use a \emph{single} outlier-aware rotation in place of the original two rotations plus permutation.
This design simultaneously halves the online rotation cost and preserves the data-dependent construction that directly targets the most problematic channels.
Furthermore, the rotation is jointly applied to the weight matrix, naturally smoothing the weight distribution.

We conduct comprehensive experiments on the LLaMA-3 model family~\citep{dubey2024llama3}, covering both pre-trained (LLaMA3-8B, LLaMA3.2-3B) and instruction-tuned (LLaMA3-8B-Instruct, LLaMA3.1-8B-Instruct) variants under MXFP4 W4A4 quantization.
As shown in Figure~\ref{fig:cross_layer_error}, the outlier-aware rotation in DuQuant++ consistently achieves the lowest per-group quantization error across all layers and positions, with the most pronounced advantage at the down projection input.
In end-to-end evaluation, DuQuant++ with GPTQ achieves a WikiText2 perplexity of 6.88 on LLaMA3-8B (FP16: 6.14) with an average zero-shot accuracy of 67.1\%, outperforming the strongest baseline MR-GPTQ by 0.41 in perplexity and 1.0\% in accuracy.
On smaller LLaMA3.2-3B, DuQuant++ reduces the perplexity from 17.95 (QuaRot) to 8.87, a relative improvement of over 50\%.

\section{Related Work}
\label{sec:related}

\subsection{Post-training Quantization}
Post-training quantization (PTQ)~\citep{ma2023ompq,ma2024outlier,yang2024dopq,yang2025lrq,lin2025quantization,lin2026efficient} has emerged as a practical paradigm for compressing large language models, as it enables efficient adaptation of pretrained networks using only a small calibration set without full retraining. 
Early studies~\citep{dettmers2022llm_int8,wei2023outlier_supp+} primarily focused on \textit{integer quantization}, beginning with INT8 and progressively moving toward more aggressive low-bit settings such as 4-bit, 3-bit, and even 2-bit representations to further reduce memory and computation costs~\citep{huang2025quaff,huang2025tequila,huang2026sherry}.
In the context of \textbf{weight-only quantization}, GPTQ~\citep{frantar2022gptq} demonstrated near-lossless INT4 compression through second-order error compensation. 
Subsequent works explored different strategies to mitigate the influence of outliers in weight matrices. 
For example, AWQ~\citep{lin2023awq} and SpQR~\citep{dettmers2023spqr} introduced alternative mechanisms for handling extreme values, while QuIP~\citep{chee2024quip}, QuIP\#~\citep{tseng2024quip_sharp}, and QTIP~\citep{tseng2024qtip} employed rotation-based transformations to regularize weight distributions and enable more aggressive compression.
For joint \textbf{weight–activation quantization}, SmoothQuant~\citep{xiao2023smoothquant} proposed redistributing quantization difficulty between weights and activations through a scaling transformation. Later approaches, such as AffineQuant~\citep{ma2024affinequant} and OmniQuant~\citep{shao2023omniquant} incorporated learnable optimization schemes to jointly refine weight and activation parameters. More recent rotation-based methods~\citep{lin2024qserve}, including QuaRot~\citep{ashkboos2024quarot}, DuQuant~\citep{lin2024duquant}, and FlatQuant~\citep{sun2024flatquant}, leverage orthogonal transformations to rebalance activation outliers, achieving strong performance under low-bit W4A4 settings.
Despite these advances, extending such techniques to floating-point microscaling formats, particularly FP4-based quantization, remains relatively underexplored.

\subsection{Microscaling Floating Point Quantization}

Microscaling floating-point formats, such as NVFP4 and MXFP4, have recently been introduced to reduce hardware and computational barriers for deploying advanced AI models, particularly with the support of Blackwell Tensor Cores~\citep{tirumala2024nvidia}. Unlike conventional uniform quantization schemes, these formats adopt block-wise fine-grained scaling, where elements within each block share a common scale factor to improve numerical efficiency.
MXFP4~\citep{rouhani2023ocp} employs a block size of 32 with scale values represented in the E8M0 format, emphasizing compact storage and efficient computation. 
NVFP4~\citep{alvarez2025introducingnvfp4} adopts a smaller group size of 16 and utilizes a full FP8 representation (E4M3) for scale encoding, allowing more precise scaling at the cost of a slightly higher bit budget per element. This design introduces a trade-off between representational accuracy and compression efficiency for both weight and activation distributions.

Recent studies~\citep{tseng2025trainingmxfp4,cook2025four,meng2026arcquant} have begun extending traditional integer-based quantization techniques to FP4 microscaling formats. 
MR-GPTQ~\citep{egiazarian2025mr-gptq} introduces block-wise Hadamard rotations combined with an efficient activation reordering strategy tailored for GPTQ, together with format-aware scale search optimizations. BRQ~\citep{shao2025brq} investigates the adaptation of standard PTQ pipelines to MXFP4 and suggests that block-wise Hadamard rotation provides a suitable transformation for this setting. MicroMix~\citep{liu2025micromix} focuses on identifying sensitive channels and preserving them at higher precision through specialized MXFP kernels.
Building upon our prior work DuQuant~\citep{lin2024duquant}, we demonstrate that fine-grained rotation can be naturally adapted to the MXFP4 format, providing an effective solution for balancing activation distributions under microscaling floating-point quantization.
\section{Preliminary}

\subsection{Integer Quantization}

Quantization converts the floating-point tensor $\mathbf{X}$ into a low-bit integer $\mathbf{X}_{q}$. Specifically, the $b$-bit uniform integer quantization can be represented as:

\begin{equation}
    \label{eq_quantization}
    \mathbf{X}_{q} = \text{clamp}\left(\left\lfloor \frac{\mathbf{X}}{s} \right\rceil \!\!+\! z, 0, 2^{b}-1 \right), 
    \textrm{where}~s=\frac{\max(\mathbf{X})-\min(\mathbf{X})}{2^b-1}, z = -\left\lfloor \frac{\min(\mathbf{X})}{s}\right\rceil. ~~~ 
\end{equation}   

The notation $\left\lfloor \cdot \right\rceil$ means the nearest rounding operation, $s$ is the quantization step size, and $z$ denotes the zero point.

\subsection{Microscaling Floating Point Format}

MXFP4 adopts a microscaling floating-point representation, where the tensor is partitioned into small blocks and each block shares a common scaling factor. 
Given a floating-point tensor $\mathbf{X}$, MXFP4 first partitions $\mathbf{X}\in\mathbb{R}^{m\times n}$ into blocks of 32 elements, denoted as $\{\mathbf{X}_j\}_{j=1}^N, N=\frac{m\cdot n}{32}$. The block-wise quantization $\mathcal{Q}(\cdot)$ for each element $\mathbf{x}_i \in \mathbf{X}_j$ is defined as:
\begin{equation}
\label{eq:quantize-mx}
\mathcal{Q}(\mathbf{x}_{i}) = \mathrm{nearest}\left(
\left\lfloor \frac{\mathbf{x}_i}{s_j} \right\rceil,
q{\min}, q{\max}
\right),
\quad
\textrm{where}~
s_j = 2^{\left\lfloor \log_2 \big( \max(|\mathbf{X}_j|) \big) \right\rfloor - b}.
\end{equation}
Here $\left\lfloor \cdot \right\rceil$ denotes rounding to the nearest representable MXFP4 value, $s_j$ is the shared block-wise scaling factor encoded in the E8M0 format, and $b$ is the format-specific exponent bias. The range $[q_{\min}, q_{\max}]$ corresponds to the valid FP4 mantissa representation.

\section{DuQuant++}
\label{sec:method}

\subsection{Motivation}

Outliers, a prominent characteristic of LLMs, are primarily determined by relatively large activation values~\citep{dettmers2022llm_int8}. These outliers are typically categorized into two types: normal outliers and massive outliers~\citep{lin2024duquant}.
Normal outliers~\citep{xiao2023smoothquant} refer to activations across all tokens with relatively large magnitudes, and they are the more prevalent type.
Massive outliers~\citep{sun2024massive,liu2024intactkv}, on the other hand, exhibit significantly larger values at a limited set of tokens. These outliers present substantial challenges for LLM quantization.

Unlike integer quantization, which typically employs per-token or per-channel scaling, the MXFP4 microscaling format partitions tensors into small blocks of 32 elements and assigns a shared E8M0 scaling factor to each block (see Eqn.~\ref{eq:quantize-mx}). This fine-grained block-wise design means that outliers within a block directly inflate the shared scaling factor $s_j$, compressing the dynamic range available for the remaining elements in that block. Consequently, the quantization error under MXFP4 is predominantly determined by the \emph{intra-block} value distribution, making it critical to reduce outlier magnitudes within each microscaling group.

\begin{figure}[t]
    \centering
    \includegraphics[width=\linewidth]{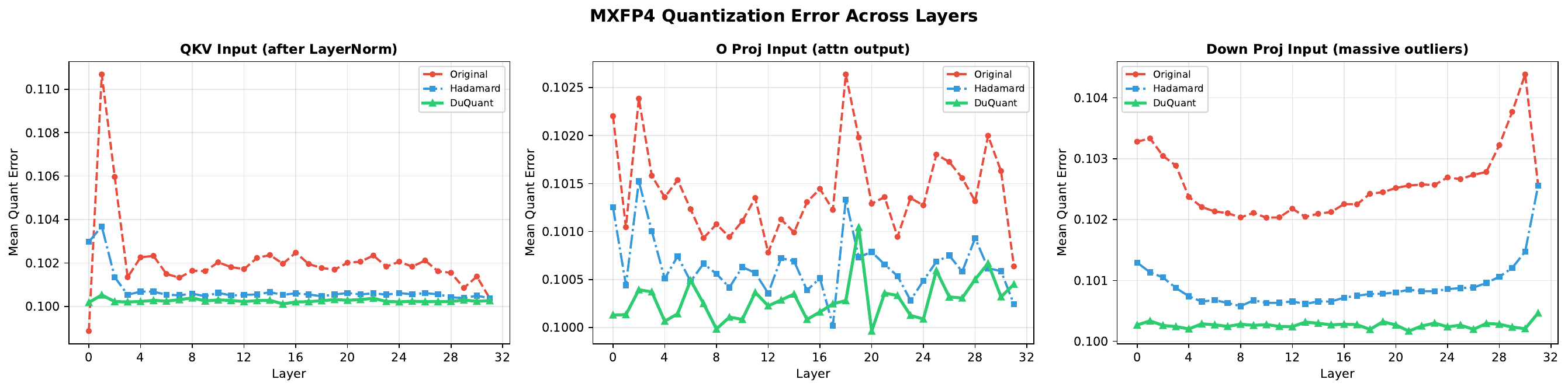}
    \caption{
    MXFP4 quantization error across all 32 layers of LLaMA-3-8B at three representative positions: QKV projection input, O projection input, and Down projection input. We compare the per-group normalized quantization error ($\|\mathbf{X}_{q} - \mathbf{X}\|_2/\|\mathbf{X}\|_2$, averaged over all groups) under three settings: the original activation (Original), block-wise randomized Hadamard rotation with block size 32 (Hadamard), and our DuQuant outlier-aware rotation with the same block size (DuQuant).
    DuQuant consistently achieves the lowest quantization error across all positions and layers, with the most significant improvement observed at the Down projection input where massive outliers exist.
    }
    \label{fig:cross_layer_error}
\end{figure}

To validate this, we conduct an empirical study on LLaMA-3-8B by measuring the MXFP4 quantization error at three key positions within each transformer layer: the QKV projection input, the output projection input, and the down projection input.
We compare three settings: (1) quantizing the original activations directly (\textit{Original}), (2) applying a block-wise randomized Hadamard rotation with block size 32, as adopted in MR-GPTQ~\citep{egiazarian2025mr-gptq} and BRQ~\citep{shao2025brq} (\textit{Hadamard}), and (3) applying the outlier-aware block-diagonal rotation from DuQuant with the same block size (\textit{DuQuant}).
As shown in Figure~\ref{fig:cross_layer_error}, the original activations exhibit substantial quantization error with high variance across layers, particularly at the down projection input where massive outliers are concentrated.
The block-wise Hadamard rotation provides a notable reduction by uniformly spreading activation energy within each group. However, Hadamard rotation still yields sub-optimal results because it is data-agnostic, where the matrix is fixed regardless of actual outlier distributions.
In contrast, DuQuant's outlier-aware fine-grained rotation, which uses the same block size but constructs the rotation matrix based on the observed outlier structure, achieves the lowest and most stable error across all positions and layers.
This result motivates our approach: under the same block-wise rotation framework aligned with the MXFP4 group size, replacing the Hadamard matrix with an outlier-aware rotation can substantially reduce the intra-group quantization error.

\subsection{DuQuant with Fine-grained Rotation}
\label{subsec:duquant_fp}

Building upon the original DuQuant method~\citep{lin2024duquant}, we adapt the rotation-based outlier mitigation framework to the MXFP4 microscaling format. The key insight is that the MXFP4 block-wise quantization structure naturally lends itself to a simplified yet effective rotation pipeline: by aligning the rotation block size with the microscaling group size, a single rotation suffices to smooth the intra-group distribution, eliminating the need for the permutation and the second rotation used in the original DuQuant.

\paragraph{Smooth Technique.}
Following SmoothQuant~\citep{xiao2023smoothquant}, we first apply a per-channel smooth transformation to shift the quantization burden from activations to weights. A diagonal scaling matrix $\mathbf{\Lambda}$ is used to rewrite the linear layer as:
\begin{equation}
    \mathbf{Y} = \mathbf{X} \cdot \mathbf{W} = (\mathbf{X} \cdot \mathbf{\Lambda}^{-1})(\mathbf{\Lambda} \cdot \mathbf{W}),
\end{equation}
where the diagonal element $\mathbf{\Lambda}_j = \text{max}(|\mathbf{X}_{j}|)^{\alpha} / \text{max}(|\mathbf{W}_{j}|)^{1-\alpha}$ and $\alpha$ controls the migration strength. This step effectively reduces normal outliers but is insufficient for massive outliers, which motivates the subsequent rotation.

\paragraph{Fine-grained Block-diagonal Rotation.}
After smoothing, we apply a single block-diagonal rotation matrix to locally redistribute the remaining outliers within each microscaling group. Crucially, we set the rotation block size $B$ to be identical to the MXFP4 group size, i.e., $B = 32$. The rotation matrix takes the form:
\begin{equation}
    \label{eq:block_rotation}
    \hat{\mathbf{R}} = \text{BlockDiag}(\hat{\mathbf{R}}_{b_1}, \hat{\mathbf{R}}_{b_2}, \ldots, \hat{\mathbf{R}}_{b_K}), \quad K = C_{in} / B,
\end{equation}
where each $\hat{\mathbf{R}}_{b_i} \in \mathbb{R}^{B \times B}$ is an orthogonal matrix constructed via the greedy outlier-aware search from DuQuant~\citep{lin2024duquant}. Specifically, the construction proceeds by: (1) identifying the feature dimension where the outlier is most concentrated, (2) building a rotation matrix that disperses the outlier energy along that dimension, and (3) iteratively repeating this process to find the step count that minimizes the peak value. Following~\citep{lin2024duquant}, all blocks share the same rotation matrix, i.e., $\hat{\mathbf{R}}_{b_i} = \hat{\mathbf{R}}_{b_k}$ for all $i$, where $b_k$ is the block containing the largest outlier. This reduces the memory cost from $K$ matrices to a single matrix.

\paragraph{Why a Single Rotation Suffices.}
In the original DuQuant designed for integer quantization, the pipeline consists of two rotations interleaved with a zigzag permutation: $\hat{\mathbf{R}}_{(1)} \to \mathbf{P} \to \hat{\mathbf{R}}_{(2)}$. The permutation is necessary because integer quantization uses per-token or per-channel scaling, where the quantization step size is determined by the global range. In this setting, block-diagonal rotation can only smooth outliers \emph{within} each block, but the \emph{cross-block} variance remains high, necessitating a permutation to redistribute outliers across blocks before applying a second rotation.

However, MXFP4 fundamentally changes this dynamic. Since each microscaling group of 32 elements has its own independent scaling factor $s_j$, the quantization error of one group does not affect another. There is no shared global scaling factor that could be ``pulled up'' by a single group's outlier. Therefore, the cross-block variance issue that motivates the zigzag permutation in the original DuQuant becomes irrelevant under MXFP4.
By setting the rotation block size equal to the MXFP4 group size ($B = 32$), each rotation block precisely corresponds to one microscaling group. A single rotation is sufficient to smooth the distribution within each group independently, and no inter-group rebalancing is required.

\paragraph{The Overall DuQuant++ Method.}
Combining the smooth technique and the fine-grained rotation, the linear layer in each transformer block is reformulated as:
\begin{equation}
    \label{eq:duquant_fp}
    \mathbf{Y} = \mathbf{X} \cdot \mathbf{W} = \underbrace{(\mathbf{X} \cdot \mathbf{\Lambda}^{-1} \cdot \hat{\mathbf{R}})}_{\hat{\mathbf{X}}} \cdot \underbrace{(\hat{\mathbf{R}}^{\top} \cdot \mathbf{\Lambda} \cdot \mathbf{W})}_{\hat{\mathbf{W}}},
\end{equation}
where the transformed activation $\hat{\mathbf{X}}$ and weight $\hat{\mathbf{W}}$ are then quantized to MXFP4 independently. Since $\hat{\mathbf{R}}$ is orthogonal, the transformation is lossless, and the inverse can be pre-absorbed into the weight matrix offline, introducing no additional overhead during inference.

\textbf{Remark 1.}
The rotation transformation is simultaneously applied to the weight matrix as $\hat{\mathbf{R}}^{\top} \cdot \mathbf{\Lambda} \cdot \mathbf{W}$. This effectively smooths the weight distribution, mitigating the outliers that the smooth technique may introduce in the weight matrix (particularly in the down-projection layer). 

\textbf{Remark 2.}
Compared to the original DuQuant, DuQuant++ uses only a single rotation with $B = 32$ instead of two rotations with a permutation. The online transformation reduces from $\mathbf{X} \to \mathbf{X} \cdot \hat{\mathbf{R}}_{(1)} \cdot \mathbf{P} \cdot \hat{\mathbf{R}}_{(2)}$ to $\mathbf{X} \to \mathbf{X} \cdot \hat{\mathbf{R}}$, halving the rotation cost and eliminating the permutation entirely. The smaller block size ($B = 32$ vs. typical $2^7$ or $2^8$ in integer quantization) further reduces the per-block matrix multiplication overhead.

\section{Experiment}

\subsection{Experimental Setup}

\paragraph{Evaluated LLMs and Quantization Baselines.}
We conduct comprehensive evaluations on three widely used large language models: the pre-trained LLaMA3-8B, and LLaMA3.2-3B, the instruction-tuned LLaMA3-8B-Instruct, and LLaMA3.1-8B-Instruct~\citep{dubey2024llama3}.
We compare against several state-of-the-art weight–activation quantization baselines, including QuaRot~\citep{ashkboos2024quarot}, FlatQuant~\citep{sun2024flatquant}, and MR-GPTQ~\citep{egiazarian2025mr-gptq}. 
For QuaRot, we consider two variants: QuaRot, which applies random rotation with RTN, and QuaRot*, which combines random rotation with GPTQ~\citep{frantar2022gptq}. 
FlatQuant~ performs the end-to-end rotation optimization tailored for quantized LLMs. 
MR-GPTQ adopts block-wise random rotation together with a revised GPTQ procedure, aiming to improve floating-point quantization performance.
Following common practice, we construct the calibration set using 128 samples drawn from WikiText2~\citep{merity2016wiki} for all baselines to ensure a fair comparison.

\paragraph{Implementation Details.}

We quantize all linear layers within the transformer blocks, following the experimental setup of MR-GPTQ~\citep{egiazarian2025mr-gptq}.
For hyperparameters, we apply a single rotation with a maximum of 128 greedy search steps.
For calibration, we randomly sample 128 sequences from the WikiText2 dataset, each with a sequence length of 2048.

\paragraph{Evaluation Benchmarks.}
We evaluate the performance of quantized LLMs using both language modeling and zero-shot question answering benchmarks.
Specifically, we report perplexity (PPL) on WikiText2~\citep{merity2016wiki} and C4~\citep{raffel2020c4}, and zero-shot accuracy on seven QA benchmarks, including ARC-E, ARC-C~\citep{clark2018arc}, HellaSwag~\citep{zellers2019hellaswag}, WinoGrande~\citep{sakaguchi2021winogrande}, LAMBADA~\citep{paperno2016lambada}, PIQA~\citep{bisk2020piqa},  and OpenBookQA~\citep{mihaylov2018openbookqa}.

\begin{table*}[!h]
\caption{
Model performance on pre-trained LLaMA3-8B and LLaMA3.2-3B with MXFP4 quantization.
}
\vspace{-10pt}
\begin{center}
\resizebox{1\linewidth}{!}{
\begin{tabular}{l|l|cc|cccccccc}
\toprule
\textbf{\#Bits}                                                                      
& \textbf{Method}      & \textbf{WikiText2} $\downarrow$ & \textbf{C4} $\downarrow$ & \textbf{HellaSwag} & \textbf{WinoGrande} & \textbf{LAMBADA} & \textbf{PIQA} & \textbf{OpenBookQA} & \textbf{ARC-E} & \textbf{ARC-C} & \textbf{Avg} $\uparrow$ \\ \midrule
\multirow{7}{*}{\textbf{\begin{tabular}[c]{@{}l@{}}LLaMA3-8B\\ MXFP4\end{tabular}}}   
& FP16                 & 6.14               & 9.46           & 79.1               & 72.9                & 75.5             & 80.7          & 44.6                & 77.6           & 53.5           & 69.1          \\ \cmidrule{2-12} 
& QuaRot               & 9.46               & 15.06          & 70.4               & 70.0                & 68.1             & 76.1          & 41.6                & 69.8           & 44.0           & 62.9          \\
& QuaRot*              & 8.07               & 13.78          & 72.3               & 68.5                & 74.6             & 76.8          & 43.2                & 75.0           & 47.3           & 65.4          \\
& FlatQuant            & 7.21               & 11.65          & 75.4               & 71.9                & 68.6             & 78.9          & 42.6                & 74.0           & 47.2           & 65.5          \\
& MR-GPTQ              & 7.29               & 11.41          & 76.4               & 69.7                & 71.8             & 77.8          & 43.4                & 74.5           & 49.4           & 66.1          \\
& \cellcolor{purple!10}\textbf{DuQuant++}  & \cellcolor{purple!10}7.07               & \cellcolor{purple!10}11.14          & \cellcolor{purple!10}76.6               & \cellcolor{purple!10}72.2                & \cellcolor{purple!10}73.0             & \cellcolor{purple!10}78.7          & \cellcolor{purple!10}41.8                & \cellcolor{purple!10}74.7           & \cellcolor{purple!10}48.2           & \cellcolor{purple!10}66.5          \\
& \cellcolor{purple!10}\textbf{DuQuant++*} & \cellcolor{purple!10}\textbf{6.88}      & \cellcolor{purple!10}\textbf{11.06} & \cellcolor{purple!10}76.6               & \cellcolor{purple!10}71.7                & \cellcolor{purple!10}74.0             & \cellcolor{purple!10}79.5          & \cellcolor{purple!10}42.6                & \cellcolor{purple!10}75.3           & \cellcolor{purple!10}50.1           & \cellcolor{purple!10}\textbf{67.1} \\ \midrule
\multirow{7}{*}{\textbf{\begin{tabular}[c]{@{}l@{}}LLaMA3.2-3B\\ MXFP4\end{tabular}}} 
& FP16                 & 7.81               & 11.34          & 74.0               & 69.5                & 69.6             & 77.7          & 40.4                & 71.8           & 46.3           & 64.2          \\ \cmidrule{2-12} 
& QuaRot               & 17.95              & 24.83          & 57.9               & 59.8                & 50.8             & 70.3          & 32.6                & 59.2           & 34.8           & 52.2          \\
& QuaRot*              & 11.46              & 18.72          & 66.1               & 63.7                & 62.2             & 73.3          & 37.6                & 62.2           & 36.6           & 57.4          \\
& FlatQuant            & 9.00               & 14.87          & 67.9               & 65.1                & 61.6             & 74.3          & 37.2                & 66.5           & 40.4           & 59.0          \\
& MR-GPTQ              & 8.79               & 13.56          & 70.1               & 65.0                & 66.6             & 76.1          & 37.8                & 70.0           & 42.8           & 61.2          \\
& \cellcolor{purple!10}\textbf{DuQuant++}  & \cellcolor{purple!10}8.87               & \cellcolor{purple!10}13.25          & \cellcolor{purple!10}70.3               & \cellcolor{purple!10}65.1                & \cellcolor{purple!10}65.3             & \cellcolor{purple!10}75.4          & \cellcolor{purple!10}39.2                & \cellcolor{purple!10}67.9           & \cellcolor{purple!10}42.9           & \cellcolor{purple!10}60.9          \\
& \cellcolor{purple!10}\textbf{DuQuant++*} & \cellcolor{purple!10}\textbf{8.63}      & \cellcolor{purple!10}\textbf{13.16} & \cellcolor{purple!10}70.6               & \cellcolor{purple!10}65.8                & \cellcolor{purple!10}67.8             & \cellcolor{purple!10}75.0          & \cellcolor{purple!10}42.2                & \cellcolor{purple!10}68.5           & \cellcolor{purple!10}42.5           & \cellcolor{purple!10}\textbf{61.8} \\ \bottomrule
\end{tabular}
}
\end{center}
\label{tab:base}
\end{table*}

\begin{table*}[!h]
\caption{
Model performance on pre-trained LLaMA3-8B-Instruct and LLaMA3.1-8B-Instruct with MXFP4 quantization.
}
\vspace{-10pt}
\begin{center}
\resizebox{1\linewidth}{!}{
\begin{tabular}{l|l|cc|cccccccc}
\toprule
\textbf{\#Bits}                                                                                  
& \textbf{Method}      & \textbf{WikiText2} $\downarrow$ & \textbf{C4} $\downarrow$    & \textbf{HellaSwag} & \textbf{WinoGrande} & \textbf{LAMBADA} & \textbf{PIQA} & \textbf{OpenBookQA} & \textbf{ARC-E} & \textbf{ARC-C} & \textbf{Avg} $\uparrow$ \\ \midrule
\multirow{7}{*}{\textbf{\begin{tabular}[c]{@{}l@{}}LLaMA3-8B\\ Instruct\\ MXFP4\end{tabular}}}   
& FP16                 & 8.31               & 13.03          & 75.7               & 71.4                & 71.5             & 78.6          & 42.8                & 78.8           & 55.6           & 67.8          \\ \cmidrule{2-12} 
& QuaRot               & 11.63              & 18.54          & 68.0               & 67.3                & 67.4             & 75.5          & 40.2                & 72.6           & 45.9           & 62.4          \\
& QuaRot*              & 10.00              & 17.43          & 69.2               & 68.8                & 71.4             & 75.4          & 39.6                & 75.8           & 48.2           & 64.1          \\
& FlatQuant            & 9.25               & 15.27          & 73.0               & 71.1                & 67.3             & 76.6          & 41.2                & 75.8           & 49.4           & 64.9          \\
& MR-GPTQ              & 9.25               & 14.62          & 73.2               & 69.0                & 68.3             & 76.0          & 43.0                & 75.8           & 51.3           & 65.2          \\
& \cellcolor{purple!10}\textbf{DuQuant++}  & \cellcolor{purple!10}8.91               & \cellcolor{purple!10}14.30          & \cellcolor{purple!10}73.8               & \cellcolor{purple!10}72.1                & \cellcolor{purple!10}69.7             & \cellcolor{purple!10}76.1          & \cellcolor{purple!10}41.2                & \cellcolor{purple!10}75.7           & \cellcolor{purple!10}52.8           & \cellcolor{purple!10}\textbf{65.9}          \\
& \cellcolor{purple!10}\textbf{DuQuant++*} & \cellcolor{purple!10}\textbf{8.75}      & \cellcolor{purple!10}\textbf{14.12} & \cellcolor{purple!10}73.7               & \cellcolor{purple!10}71.6                & \cellcolor{purple!10}70.3             & \cellcolor{purple!10}76.7          & \cellcolor{purple!10}41.2                & \cellcolor{purple!10}77.2           & \cellcolor{purple!10}50.7           & \cellcolor{purple!10}\textbf{65.9} \\ \midrule
\multirow{7}{*}{\textbf{\begin{tabular}[c]{@{}l@{}}LLaMA3.1-8B\\ Instruct\\ MXFP4\end{tabular}}} 
& FP16                 & 7.21               & 11.39          & 79.2               & 74.1                & 73.2             & 80.9          & 43.0                & 79.6           & 55.2           & 69.3          \\ \cmidrule{2-12} 
& QuaRot               & 10.42              & 16.72          & 72.4               & 68.4                & 62.0             & 76.0          & 39.8                & 71.8           & 46.3           & 62.4          \\
& QuaRot*              & 9.23               & 16.16          & 71.8               & 70.7                & 72.5             & 76.9          & 39.4                & 76.3           & 51.9           & 65.6          \\
& FlatQuant            & 8.10               & 13.40          & 75.7               & 71.1                & 69.6             & 78.7          & 41.8                & 78.3           & 52.1           & 66.8          \\
& MR-GPTQ              & 9.06               & 12.94          & 76.2               & 70.6                & 70.4             & 78.4          & 43.4                & 77.2           & 52.6           & 67.0          \\
& \cellcolor{purple!10}\textbf{DuQuant++}  & \cellcolor{purple!10}8.03               & \cellcolor{purple!10}12.96          & \cellcolor{purple!10}77.3               & \cellcolor{purple!10}70.6                & \cellcolor{purple!10}70.2             & \cellcolor{purple!10}79.3          & \cellcolor{purple!10}42.0                & \cellcolor{purple!10}78.0           & \cellcolor{purple!10}52.1           & \cellcolor{purple!10}67.1          \\
& \cellcolor{purple!10}\textbf{DuQuant++*} & \cellcolor{purple!10}\textbf{7.89}      & \cellcolor{purple!10}\textbf{12.89} & \cellcolor{purple!10}77.0               & \cellcolor{purple!10}71.7                & \cellcolor{purple!10}72.3             & \cellcolor{purple!10}79.5          & \cellcolor{purple!10}42.4                & \cellcolor{purple!10}77.0           & \cellcolor{purple!10}51.7           & \cellcolor{purple!10}\textbf{67.4} \\ \bottomrule
\end{tabular}
}
\end{center}
\label{tab:instruct}
\end{table*}

\subsection{Main Results}

We present the MXFP4 quantization results on pre-trained models in Table~\ref{tab:base} and instruction-tuned models in Table~\ref{tab:instruct}. Our key findings are summarized as follows.

\paragraph{DuQuant++ consistently achieves the best overall performance.}
Across all four evaluated models, DuQuant++ and DuQuant++* consistently outperform existing baselines in terms of both perplexity and average zero-shot accuracy.
On LLaMA3-8B, DuQuant++* achieves a WikiText2 perplexity of 6.88 and an average accuracy of 67.1\%, narrowing the gap to the FP16 baseline to only 0.74 in perplexity and 2.0\% in accuracy.
In contrast, the strongest competing method, MR-GPTQ, attains a perplexity of 7.29 and an average accuracy of 66.1\%, lagging behind DuQuant++* by 0.41 in perplexity and 1.0\% in accuracy.
Similarly, on the smaller LLaMA3.2-3B model, DuQuant++* achieves the lowest C4 perplexity (13.16) and the highest average accuracy (61.8\%), surpassing MR-GPTQ by 0.6\% in average accuracy, which demonstrates the scalability of our approach to smaller model sizes.

\paragraph{Fine-grained rotation is more effective than global rotation for MXFP4.}
A consistent observation across all settings is that DuQuant++ substantially outperforms QuaRot, which applies global random Hadamard rotation.
For instance, on LLaMA3.2-3B, QuaRot suffers a severe perplexity degradation (17.95 on WikiText2), while DuQuant++ reduces it to 8.87---a relative improvement of over 50\%.
This stark contrast highlights that global rotation alone is insufficient for the MXFP4 format, where the shared exponent within each microscaling block amplifies the impact of outlier activations.
By operating at a finer granularity, DuQuant++ more effectively redistributes outlier magnitudes across channels, yielding a smoother activation distribution that is significantly more amenable to microscaling quantization.
Compared with FlatQuant, which performs end-to-end rotation optimization, DuQuant++ achieves competitive or superior results without requiring costly learnable rotation matrices, demonstrating the efficiency of our fine-grained rotation strategy.

\paragraph{GPTQ provides complementary benefits.}
Comparing DuQuant++ and DuQuant++* across all models, we observe that incorporating GPTQ consistently improves both perplexity and accuracy.
On LLaMA3-8B, adding GPTQ reduces WikiText2 perplexity from 7.07 to 6.88 and boosts the average accuracy from 66.5\% to 67.1\%.
This improvement is also evident on instruction-tuned models: on LLaMA3.1-8B-Instruct, DuQuant++* achieves a perplexity of 7.89 versus 8.03 for DuQuant++, with average accuracy increasing from 67.1\% to 67.4\%.
These results confirm that our fine-grained rotation and second-order weight compensation are complementary, the rotation smooths activation outliers to facilitate quantization, while GPTQ further minimizes the weight quantization error.

\paragraph{Generalization to instruction-tuned models.}
As shown in Table~\ref{tab:instruct}, DuQuant++ and DuQuant++* maintain their superiority on instruction-tuned variants.
On LLaMA3-8B-Instruct, DuQuant++* achieves the best perplexity (8.75 on WikiText2) and the highest average accuracy (65.9\%), outperforming all baselines.
Notably, on LLaMA3.1-8B-Instruct, DuQuant++* attains a WikiText2 perplexity of only 7.89, which is remarkably close to the FP16 baseline of 7.21, while maintaining an average accuracy of 67.4\%, only 1.9\% below full precision.
This suggests that the fine-grained rotation learned from calibration data generalizes well across different training paradigms, making DuQuant++ a robust and practical solution for deploying quantized LLMs in real-world applications.

\section{Conclusion}

We presented DuQuant++, a simple yet effective approach to MXFP4 weight-activation quantization for large language models.
By aligning the outlier-aware block-diagonal rotation with the MXFP4 microscaling group size, DuQuant++ exploits the independent scaling structure of MXFP4 to collapse the dual-rotation-plus-permutation pipeline of the original DuQuant into a single rotation, halving the online transformation cost while preserving data-dependent outlier suppression.
Comprehensive experiments on four LLaMA-3 models demonstrate that DuQuant++ consistently achieves state-of-the-art MXFP4 W4A4 performance, narrowing the gap to full-precision models.
We hope our findings encourage further exploration of format-aware rotation design for emerging low-bit floating-point quantization schemes.


\newpage

\bibliography{nips}
\bibliographystyle{plainnat}

\end{document}